# Integration of Explainable AI Techniques with Large Language Models for Enhanced Interpretability for Sentiment Analysis

**Thivya Thogesan[1], Anupiya Nugaliyadde[1,2], Kok Wai Wong[1]**

[1]School of Information Technology, Murdoch University
[2]Norwood Systems
thivya.thogesan@murdoch.edu.au[*]
a.nugaliyadde@murdoch.edu.au, anupiya.nugaliyadde@norwoodsystems.com
k.wong@murdoch.edu.au

**Abstract**

Interpretability remains a key difficulty in sentiment analysis with Large Language Models (LLMs), particularly in high-stakes applications where it is crucial to comprehend the rationale behind forecasts. This research addressed this by introducing a technique that applies SHAP (Shapley Additive Explanations) by breaking down LLMs into components such as embedding layer, encoder, decoder and attention layer to provide a layer-by-layer knowledge of sentiment prediction. The approach offers a clearer overview of how model interpret and categorise sentiment by breaking down LLMs into these parts. The method is evaluated using the Stanford Sentiment Treebank (SST-2) dataset, which shows how different sentences affect different layers. The effectiveness of layer-wise SHAP analysis in clarifying sentiment-specific token attributions is demonstrated by experimental evaluations, which provide a notable enhancement over current whole-model explainability techniques. These results highlight how the suggested approach could improve the reliability and transparency of LLM-based sentiment analysis in crucial applications.

## Introduction

Sentiment analysis, or opinion mining, is a crucial task in natural language processing (NLP) that aims to identify and categorize opinions expressed in text, particularly to determine whether the sentiment is positive, negative, or neutral (Pang and Lee 2008). Over the last decade, Large Language Models (LLMs) such as GPT (Radford et al. 2018), BERT (Devlin et al. 2019), and their successors have revolutionized sentiment analysis by significantly improving the accuracy and contextual understanding of textual data. These models leverage pre-training on massive datasets and deep learning architectures to capture subtle nuances in language, allowing them to handle complex sentence structures and context more effectively than traditional methods (Zhang, Wang, and Liu 2021). As a result, LLMs have set new performance benchmarks in sentiment analysis tasks, enabling more accurate classification of sentiments, even in difficult cases like sarcasm or irony (Sun, Huang, and Qiu 2019).

Despite these advancements, the interpretability of LLMs remains a major concern. Even while LLMs perform at the cutting edge of sentiment analysis, they frequently function as "black boxes," offering little information about how they make certain sentiment predictions (Rudin 2019). This opacity is problematic in domains where understanding the reasoning behind a sentiment prediction is critical, such as healthcare, finance, or legal applications. In these sectors, decisions based on sentiment analysis need to be explainable to ensure that the models' predictions can be trusted and validated (Doshi-Velez and Kim 2017a). In business, for example, companies rely on sentiment analysis to gauge customer feedback, monitor brand perception, and make strategic decisions (Zhang, Wang, and Liu 2021). A model that predicts customer sentiment but cannot explain why it reached that prediction can lead to misinterpretation and potentially costly decisions. In healthcare, sentiment analysis is increasingly used to assess patient feedback and emotional states, influencing healthcare strategies and patient outcomes. Here, a lack of transparency in the model's decisions can have serious consequences, particularly if the sentiment analysis informs clinical judgments or patient care (Gunning and Aha 2019).

Traditional explainability methods, such as attention mechanisms (Bahdanau, Cho, and Bengio 2015) and gradient-based methods (Sundararajan, Taly, and Yan 2017), offer limited insights into how Large Language Models (LLMs) make predictions by focusing mainly on surface-level features, like output and attention layers. These methods often fail to provide deeper, phrase-level understanding or explain the internal transformations that occur in layers like embeddings and encoders (Jain and Wallace 2019). Recent advancements, such as SHAP and LIME (Lundberg and Lee 2017; Ribeiro, Singh, and Guestrin 2016), have improved interpretability by identifying key input features (Mackova, Wang, and Mitra 2021) but still lack comprehensive coverage of LLMs' multi-layered structures. Most of the existing research has focused on dissecting LLMs into two main components, namely multi-head self-attention (MHSA) and multi-layer perceptron (MLP), there is potential for further decomposition of LLMs into additional layers.

This research seeks to bridge the gap in interpretability by systematically analyzing the individual components of

Large Language Models (LLMs). By applying Shapley Additive Explanations (SHAP) to key layers—embedding, encoder, and attention layers. It aims to uncover how these elements contribute to the model's behavior and predictions, with a specific focus on phrase-level analysis.

A novel approach is proposed that applies SHAP to pretrained BERT models, leveraging the Stanford Sentiment Treebank (SST-2) (Socher et al. 2013) and IMDB dataset for validation(Maas et al. 2011). The proposed approach was compared to an existing method that applied SHAP directly to the BERT model's outputs. Evaluation results demonstrated that the proposed framework provided more granular insights into the model's decision-making process, outperforming the existing approach in enhancing transparency and trustworthiness in sentiment classification. This advancement paves the way for more interpretable and accountable AI systems, particularly in high-stakes applications.

# Literature Review

Explainability, defined by (Doshi-Velez and Kim 2017b), refers to the ability of AI systems, including LLMs, to give clear, transparent explanations for their decisions and predictions. It essentially aims to "open the black box" of complex models so that practitioners, academics, and end-users can comprehend how the model arrived at a particular outcome. Explainability is crucial for fostering confidence in model predictions and ensuring responsible deployment in high-stakes industries like healthcare, finance, and law, where trust, responsibility, and compliance are paramount.

In response to the opacity of LLMs, several Explainable AI (XAI) approaches have been developed. These techniques aim to shed light on how models such as BERT and GPT handle input data, generate predictions, and prioritize different features. The main XAI methods for elucidating LLM predictions are examined below, with an emphasis on their use in sentiment analysis.

## SHapley Additive exPlanations (SHAP)

SHAP is a widely recognized game-theoretic approach that explains model predictions by assigning importance values to individual input features (Lundberg and Lee 2020). SHAP has been extensively used in sentiment analysis to identify which tokens most influence predictions in models like BERT. (Barnhart and Balakumar 2020) applied SHAP to explain BERT models at the input embedding layer, identifying key tokens that influenced sentiment predictions. (Mackováˊ, Šimko, and Bielikováˊ 2021) extended SHAP to explore the attention heads in transformer models, focusing on the internal components that modulate token importance across layers. However, early versions of SHAP did not adequately explain intermediate layers like attention mechanisms or encoder layers. The introduction of TransSHAP by (Chen, Zhang, and Sun 2023) addressed this limitation by assigning SHAP values not only to tokens but also to intermediate components like attention heads and encoder layers, offering a complete layer-wise analysis.

## Local Interpretable Model-agnostic Explanations (LIME)

LIME is a model-agnostic XAI technique that provides localized explanations by perturbing the input data and using simpler models to approximate the behavior of complex models (Ribeiro, Singh, and Guestrin 2016). LIME has been applied to sentiment analysis tasks to highlight key words driving predictions. However, LIME generally focuses on the input layer, perturbing tokens to understand their impact on the output. (Narayan and Verma 2021) applied LIME to transformers in the healthcare domain but restricted their analysis to token-level explanations, not delving into deeper model layers. LIME's lack of focus on intermediate layers like attention heads or hidden layers limits its ability to fully capture how transformers make decisions.

## Attention Visualization and Attribution

Attention mechanisms are integral to transformers like BERT and GPT, and various visualization techniques have been developed to interpret how models utilize attention weights. (Jain and Wallace 2019) analyzed attention scores at the attention head level and found that these scores do not always correlate with the true importance of features. (Abnar and Zuidema 2021) improved upon this by proposing hierarchical attention attribution, analyzing attention weights across multiple transformer layers. However, attention visualization techniques often fail to explain other components like embedding layers or the interactions between layers.

## Integrated Gradients

Integrated Gradients, a gradient-based method, calculates feature attributions by measuring how model predictions change as the input transitions from a baseline to its actual value (Sundararajan, Taly, and Yan 2017). (Chefer, Gur, and Wolf 2020) applied Integrated Gradients to transformer models, focusing on token-level attributions and tracking how individual tokens affect predictions as they propagate through layers. However, Integrated Gradients typically do not explain intermediate components like attention heads or encoder layers, limiting the depth of layer-wise insights.

## Rationale Extraction

Rationale extraction aims to provide human-understandable explanations by selecting the most important parts of the input that directly influence the model's decision (Lei, Barzilay, and Jaakkola 2016). While rationale extraction offers token-level explanations, it does not account for how internal components like attention heads or hidden layers contribute to decisions, leaving gaps in understanding the full decision-making process in transformer models.

## Layer-wise Relevance Propagation (LRP)

Layer-wise Relevance Propagation (LRP) assigns relevance scores to input features and propagates these scores backward through the model's layers (Arras et al. 2020). LRP has been applied to BERT models, providing insights into how tokens contribute to predictions across layers. However, LRP's computational expense makes it difficult to scale to larger models like GPT, and it does not focus on specific components like attention heads or encoder layers, limiting its ability to provide detailed explanations of how transformers make decisions.

This research addresses key gaps in existing XAI methods by offering a more granular, component-specific explanation of how LLMs like BERT and GPT make decisions in sentiment analysis. While current approaches such as LIME (Ribeiro, Singh, and Guestrin 2016), focus on input token perturbations but do not explore deeper layers, while attention visualization techniques (Jain and Wallace 2019; Abnar and Zuidema 2021) are limited to understanding attention mechanisms without considering other components of the model architecture. Integrated Gradients (Sundararajan, Taly, and Yan 2017) and Layer-wise Relevance Propagation (Arras et al. 2020) similarly focus on token-level attributions or layer-wise relevance, but they do not offer detailed insights into the role of individual components across the model.

This research introduces a SHAP-based methodology that applies to each individual component of LLMs, offering a detailed, layer-wise breakdown of the model's internal processes. By isolating and explaining each part of the architecture, it provides more precise explanations for sentiment classification, enhancing interpretability and transparency compared to existing methods.

## Methodology

This research utilized a pretrained BERT model, fine-tuned on the Stanford Sentiment Treebank (SST) dataset, with SHAP as the explainability technique. The SST dataset was chosen for its fine-grained sentiment annotations, enabling detailed analysis of interpretability at various levels (Socher et al. 2013). BERT, with its efficient self-attention-based architecture and pretraining on large corpora, was selected for its state-of-the-art performance in NLP tasks. SHAP, grounded in cooperative game theory, provided fair and consistent feature attributions, offering both local and global interpretability. Its ability to attribute importance at the phrase level made it particularly suitable for rigorous sentiment analysis (Lundberg and Lee 2017; Zhao, Chen, and Yang 2024).

### Proposed Explainability Approach

The proposed approach, shown in Figure 1, is to break down the LLM into its most basic parts: an output layer, an encoder, a decoder, and an embedding layer. Each component uses explainable Artificial Intelligence (XAI) methodologies to identify words and qualities that have the most influence on the sentiment analysis findings the component generates. The major objective is to integrate these insights from disparate components into a cohesive framework that offers comprehensive insight into the model's decision-making process. The words that are highlighted and the explanations derived from each component can be combined to make graphs. These visualisations effectively highlight the emphasis and contribution of each word across several components, transforming the model into a "white box."

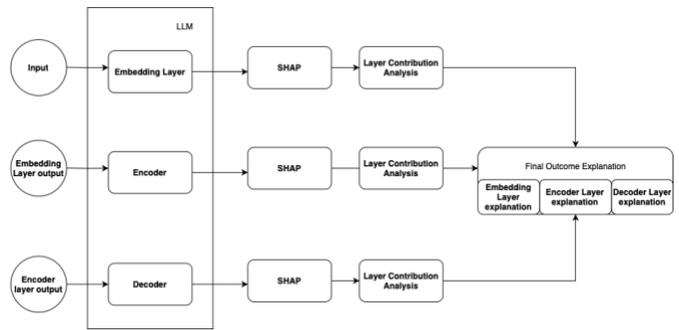

Figure 1: Architecture diagram

This research employs SHAP (Shapley Additive Explanations) to explain the decision-making process of a fine-tuned BERT model used for sentiment analysis on the SST-2 dataset. The model is analyzed at multiple levels: embedding, encoder, and attention as shown in detailed Figure 2.BERT was selected to be used in this research, Therefore SHAP explainers are applied to both the embedding and encoder layers only. However, this approach can be extended to include decoders in other large language models, such as those used in models like GPT.SHAP values are computed to measure the contribution of each token or phrase to the model's sentiment predictions. This process helps reveal how the model assigns importance to different phrases across layers.

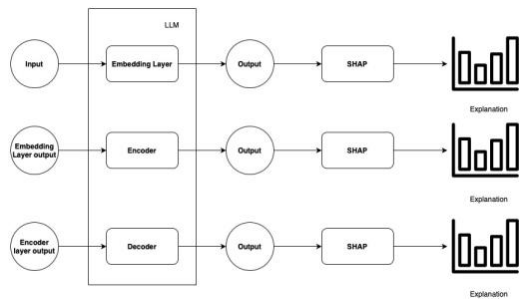

Figure 2: Layer wise explainability

The SHAP value calculation process is done as shown in Figure 2.

First, the input sentence is tokenized using the BERT tokenizer, followed by phrase extraction using spaCy's linguistic parser. This process segments the text into meaningful components, including noun phrases, verb phrases, adjective phrases, and prepositional phrases, ensuring that SHAP values can be attributed not only to individual tokens but also to entire phrases.The next step involves phrase masking, where a custom masker replaces entire phrases with padding tokens ("[PAD]"), simulating their absence. This is fundamental to the SHAP method, which perturbs the input to observe how the model's output changes. By masking phrases rather than individual tokens, the method captures the impact of entire phrases on the sentiment classification.

The SHAP explainer is then initialized, passing masked inputs to the BERT model through a lambda function, which allows us to compute SHAP values at two critical layers: the embedding and encoder layers. In the embedding layer, each token is transformed into a dense vector representing its semantic properties. The encoder layer applies self-attention mechanisms to capture relationships between tokens, generating context-sensitive representations. For each masked input, the SHAP explainer calculates how much removing the phrase changes the model's output, thereby assigning importance scores to the phrases.

To scientifically represent the aggregation of SHAP values by phrase, The following notation can be derived:

$$\Phi(p_i) = \sum_{w_j \in p_i} \phi_i(w_j)$$

Where $\Phi(p_i)$ is the aggregated SHAP value for phrase $p_i$, $w_j$ is a word in $p_i$, $\phi_i(w_j)$ is the SHAP value for $w_j$ in the context of $p_i$, and $\sum_{w_j \in p_i}$ denotes summation over all words in the phrase.

For each phrase in the input, SHAP calculates the difference in the model's prediction with and without the phrase present. Positive SHAP Values: Indicate that the phrase contributes towards the model's confidence in predicting a positive sentiment. Negative SHAP Values: Suggest that the phrase contributes towards a negative sentiment prediction.

The SHAP values which were calculated at both the embedding and encoder layers are aggregated. This aggregation is done across layers (embedding and encoder) to provide a more holistic understanding of the contribution of each phrase and it can be understood by the below equation:

$$AggregatedSHAP = \sum_{l=1}^{L} SHAP_l$$

Where:
- $L$ is the number of layers (embedding, encoder, etc.).
- $SHAP_l$ represents the SHAP values from the $l$-th layer.

This aggregation provides insight into the overall importance of a phrase by combining the contributions of its constituent words, highlighting how these phrases influence sentiment predictions across different layers of the model. The visualization of SHAP values, where positive contributions are shown in green and negative contributions in red, enables a clearer understanding of how individual words and phrases impact the model's decision-making process.

By applying these methods, the internal workings of the BERT model are made transparent, allowing for better interpretation of how it arrives at specific predictions.

## Experimentation

The experimentation involved testing and genertating explanation on all the sentences from both IMDB Dataset (Maas et al. 2011) and SST dataset. The sentences were tokenized, and key phrases were extracted using spaCy's linguistic parsing capabilities. SHAP values were calculated for the embedding layer, encoder layer, and output layer of the model. By aggregating SHAP values across layers, a holistic view of the model's reasoning process was obtained, revealing how the model derived its output. Additionally, attention weights were analyzed, providing insights into how the model distributed focus across tokens, with particular emphasis on key phrases. Experimentation was conducted with various types of sentences, including those with a sarcastic tone.

As discussed in previous works, SHAP has been applied directly to large language model outputs for explanations. This research adopts (Kokalj et al. 2021)'s methodology as a benchmark for evaluation. SHAP was applied to the BERT model while keeping all other factors constant. This baseline approach operates at the token level, as explored in existing work (Kokalj et al. 2021).

## Results, Discussion and Evaluation
### Proposed Explainability Approach

Results of following sentences obtained from the two datasets will be explored.

**S1**:*"neither parker nor donovan is a typical romantic lead , but they bring a fresh , quirky charm to the formula ."*

**S2**:*"Read the book, forget the movie!"*

**S3**:*"Oh great, another email. I just love waiting an extra week for something I ordered two months ago."*

**Embedding Layer** :

The SHAP values at the embedding layer reveal how individual phrase meanings, contextual relationships, and patterns from training data interact.

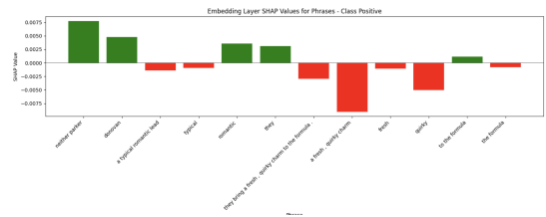

Figure 3: embedding layer wise explainability for S2

For Sentence 1 (S1) in Figure 3, the phrase *"neither parker"* carries a positive SHAP value for establishing neutrality, and *"donovan"* contributes similarly due to its isolated neutral tone. However, *"a typical romantic lead"* has a negative SHAP value overall because while *"romantic"* is positive, *"typical"* conveys critique, shifting the sentiment negatively. The word *"they"* has a positive SHAP value, serving as a neutral-to-positive transition.

Phrases like *"a fresh, quirky charm"* carry negative SHAP values despite positive words like *"fresh"* and *"quirky,"* as the embedding layer associates these descriptors with critiques in context. Similarly, *"to the formula"* and *"the formula"* have negative SHAP values, indicating conformity and lack of originality, common in critiques.

The SHAP values stem from factors such as each phrase's inherent sentiment, sentence position, and interactions with surrounding phrases. For instance, positive terms like *"romantic"* are influenced by prior critique-heavy contexts, diminishing their impact. Transitions like *"but"* also retroactively shift sentiment. Patterns learned from the SST-2 training data further shape how phrases with specific structures or transitions are encoded as critiques, explaining the embedding layer's sentiment interpretations.

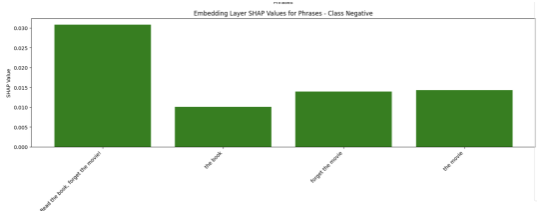

Figure 4: embedding layer wise explainability for S2

The SHAP values for the sentence 2 (s2) as shown in Figure 4 highlight how analyzing the entire phrase first impacts the interpretation of its subphrases.

The complete phrase has the strongest contribution to the negative class sentiment, setting a critical tone that frames subsequent analysis. Subphrases like *"the book"* inherit a favorable sentiment due to its positive comparison to the movie, while *"forget the movie"* intensifies the critique by reinforcing the dismissal of the movie. The embedding layer uses patterns from the whole-phrase analysis to guide the sentiment attribution of each subcomponent, ensuring their contributions align with the overarching sentiment.

**Encoder Layer** At the encoder layer as shown in Figure 5, SHAP values are derived by analyzing both the inherent meanings of phrases and their relationships within the broader context of the sentence. For instance, the phrase *"neither parker"* receives a positive SHAP value as the encoder identifies it as part of a neutral, balanced comparison, setting up a tone for subsequent contrasts. Similarly, *"donovan"* contributes positively due to its association with the neutral tone established earlier. However, the phrase *"a typical romantic lead"* introduces subtle negativity, with the encoder recognizing the word *"typical"* as a critique that diminishes the positivity of *"romantic lead."* While *"romantic"* retains its inherent positive SHAP value at the encoder layer, it is somewhat overshadowed by the critique implied by *"typical."*

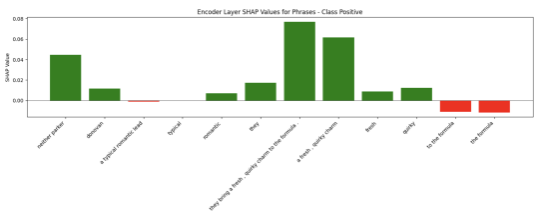

Figure 5: encoder layer wise explainability.

As the sentence progresses, *"they"* continues to contribute positively, maintaining neutrality and facilitating a transition in focus. Phrases like *"they bring a fresh, quirky charm to the formula"* are strongly positive in the encoder layer, with words like *"fresh"* and *"charm"* recognized as impactful sentiment carriers. However, the encoder dynamically processes transitions, and when phrases like *"to the formula"* and *"the formula"* appear, they signal conformity or rigidity, which detracts significantly from the earlier positive sentiment, resulting in strong negative SHAP values.

This layer-level analysis reveals how phrases interact contextually. For example, the encoder amplifies the positive sentiment of *"a fresh, quirky charm"* by linking it with the earlier neutral-to-positive shift, but the contrast introduced by *"to the formula"* diminishes this positivity, shifting the sentiment trajectory. The encoder layer processes these sequential dependencies, dynamically adjusting SHAP values to reflect shifts in tone or meaning.

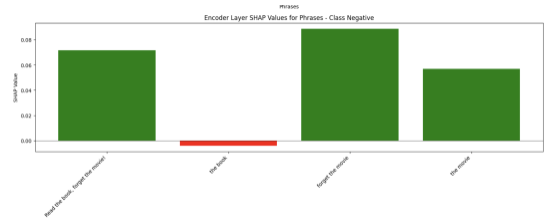

Figure 6: encoder layer wise explainability for S2

For S2, as shown in Figure 6, reveals that *"Read the book"* positively influences sentiment, while *"the book"* alone introduces a negative sentiment. *"Forget the movie"* strongly reinforces a positive sentiment, and *"the movie"* has a moderate positive effect. This analysis shows how the encoder assigns varying positive and negative SHAP values to phrases, capturing a nuanced sentiment balance across the sentence.

**Attention Layer** The model's attention analysis shows how key phrases shape sentiment.

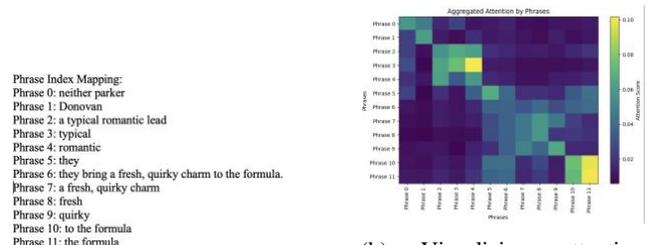

(a) Mapping for phrases.  (b) Visualizing attention scores by phrases.

Figure 7: Mapping and attention scores for s1

In relevant to figure 7, Significant attention is given to *"typical"* and *"fresh"*, highlighting their roles in altering sentiment, while phrases with minimal overlap, like *"neither parker"* and *"to the formula"*, receive less focus. Longer sentiment-rich phrases, such as *"they bring a fresh, quirky charm to the formula"*, are independently significant. Overall, the model prioritizes crucial sentiment-driving phrases

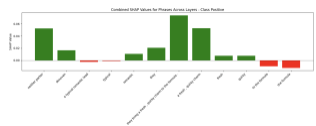
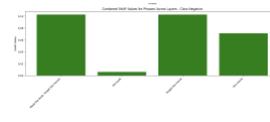

(a) Aggregated analysis for S1

(b) Aggregated analysis for S2

Figure 9: Aggregated Analysis for S1 and S2.

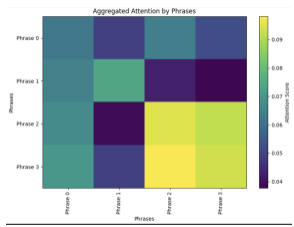

(a) Mapping for phrases.

(b) Visualizing attention scores by phrases.

Figure 8: Mapping and attention scores.

For s2 as shown in Figure 8, Phrases like "the book" and "forget the movie" receive notable attention, reflecting their relevance in shaping the sentiment. In contrast, shorter or less sentiment-rich phrases, such as "the movie", are assigned lower attention scores. The heatmap demonstrates the model's ability to focus on meaningful phrases, suggesting that it captures nuanced relationships between phrase-level sentiment contributions. This indicates a structured approach to interpreting sentiment, with emphasis on phrases that drive overall sentiment shifts.

**Aggregated Layer Analysis** The aggregated SHAP analysis provides a comprehensive view of how phrases contribute to sentiment prediction by integrating insights from both the embedding and encoder layers. In sentence 1 as shown in Figure 9a , Neutral phrases like *"neither parker"* and *"donovan"* show positive SHAP values due to their role in maintaining balance. Negative sentiment emerges with phrases such as *"a typical romantic lead,"* where the embedding layer identifies *"typical"* as a critique, and the encoder layer amplifies this contextually. Positive contributions, such as *"fresh"* and *"charm,"* are strongly emphasized, particularly by the encoder layer, while transitional phrases like *"to the formula"* and *"the formula"* introduce strong negative SHAP values, marking a sentiment shift toward criticism. This analysis highlights the interplay of positive and negative contributions, with the embedding layer focusing on local meanings and the encoder layer contextualizing them, showcasing the dynamic evolution of sentiment across layers.

For s2 as shown in Figure 9b, The analysis illustrates how phrases evolve across layers to contribute to the final sentiment outcome. Neutral terms like *"the book"* and *"forget the movie"* initially carry balanced SHAP values in the embedding layer, reflecting their foundational significance. Negative phrases such as *"the book"* gain prominence as the encoder layer contextualizes their inefficiencies, while positive phrases like *"forget the movie"* and *"the mvoie"* show increasingly strong SHAP values, highlighting their transformative potential.

During experimentation with a sarcastic tone, the model incorrectly classified the sentiment as positive. For example, the sentence *"Oh great, another email. I just love waiting an extra week for something I ordered two months ago."* was misinterpreted, with phrases such as *"Oh great, another email"* and *"I just love waiting"* contributing to a positive sentiment prediction, likely due to the literal interpretation of words like *"great"* and *"love"*.

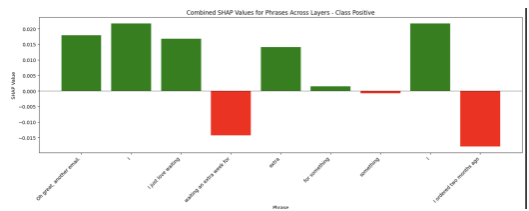

Figure 10: visualizing shap values for all layers.

However, the SHAP analysis provided key insights into the model's decision-making process. The SHAP values as shown in Figure 9 showed that phrases like *"waiting an extra week"* and *"I ordered two months ago"* had negative SHAP values, correctly contributing to the negative sentiment classification. This demonstrates how SHAP helps to identify where the model misinterprets sentiment, particularly in cases of sarcasm, by showing how different parts of the input text influence the prediction.

**Evaluating Against Baseline Model**

Human evaluation was excluded due to the time constraints and complexity involved in organizing and analyzing subjective assessments.The decision to avoid validating against other XAI techniques ensured methodological focus and consistency, avoiding variability and complexity that could detract from the primary objective. Figure 11 shows the result obtained for the baseline approach which is to apply SHAP to the BERT model's output while keeping all other factors the same for S1.

The positive SHAP values for tokens such as *"neither"*, *"parker"*, and *"romantic"* highlight their strong influence in reinforcing the model's positive sentiment prediction. The word *"quirky"* is split into *"qui"* and *"rky"* due to BERT's subword tokenization. This tokenization can cause interpretability issues, as the SHAP values are now assigned to parts of a word rather than the full word itself. Conversely, some tokens like *"bring"*, *"they"*, and *"but"* exhibit negative SHAP values, suggesting that they detract from the positive sentiment classification.

The above benchmark will be used to evaluate the proposed explainability at different component levels of the model.

The baseline SHAP analysis as shwon in Figure 11 provides a static view of token contributions by attributing sentiment values directly to the final output without considering how these values are shaped across layers. While it identifies words like *"romantic"* and *"lead"* as strongly positive and *"but"* and *"the formula"* as strongly negative, it lacks the ability to capture contextual interactions, transitions, or relationships between phrases. Additionally, subword tokenization issues, such as splitting *"quirky"* into *"qui"* and *"rky,"* further limit interpretability. In contrast, the embedding layer as shown in Figure 3 improves on the baseline by capturing the initial semantic meanings of tokens and their immediate relationships. For instance, it recognizes *"romantic"* as inherently positive and *"typical"* as slightly negative, reflecting learned patterns from training data. However, while the embedding layer provides early-stage insights, it does not fully contextualize contributions or capture long-range dependencies.

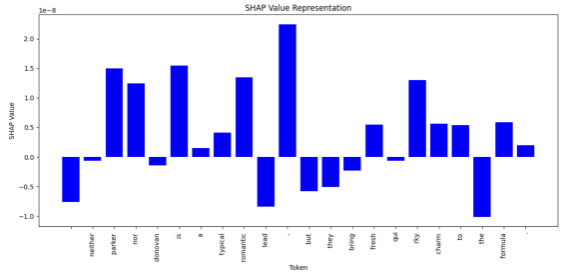

Figure 11: Visualizing SHAP values for the baseline model.

The encoder layer as shown in Figure 5 builds on this by introducing contextual relationships and interactions between tokens. It highlights how context amplifies or diminishes contributions, such as *"romantic"* retaining positivity but being reduced within *"a typical romantic lead"* due to the critique implied by *"typical."* The encoder layer also emphasizes transitions like *"but,"* showing how they redirect sentiment by linking earlier positive phrases like *"a fresh, quirky charm"* to later negative ones like *"to the formula."* Unlike the baseline as shown in Figure 3, which treats these transitions in isolation, the encoder layer reveals their critical role in sentiment shifts. Furthermore, it captures relationships between phrases, such as how *"to the formula"* and *"the formula"* collectively drive a strong negative sentiment, which the baseline fails to explain.

The aggregated analysis as shown in Figure 9 offers a holistic view that significantly surpasses the baseline. It traces how contributions evolve from isolated token semantics in the embedding layer to contextual relationships in the encoder layer. For instance, it resolves subword tokenization issues by combining fragmented tokens like *"quirky"* into meaningful phrases and highlights how positive phrases like *"a fresh, quirky charm"* initially dominate sentiment but are later overshadowed by negative shifts caused by *"to the formula"* and *"the formula."* Unlike the baseline as shown in Figure 11, which provides static SHAP values, the aggregated analysis explains how and why these contributions develop, offering comprehensive interpretability of sentiment predictions.

For sarcastic and incorrect predictions, the baseline fails to trace why certain tokens contribute disproportionately to the sentiment classification. The proposed approach result for sarcasm as shown in Figure 13a is the aggregated view, and it can provide detailed insights into how and why the model's predictions evolve through embedding layer as shown in Figure 12a and encoder layer as shown in Figure 12b. For instance, when certain tokens or phrases (e.g., "but") cause a sentiment shift, The proposed approach identifies their contextual influence through attention and encoder layer analysis. By tracking SHAP values across layers, the proposed method can pinpoint phrases that drive misclassification, offering actionable insights to refine the model or training data.

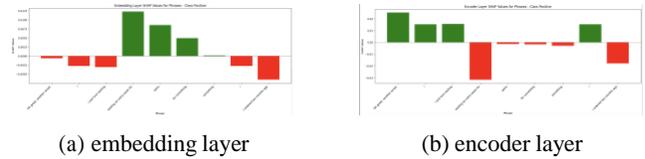

(a) embedding layer        (b) encoder layer

Figure 12: layer wise analysis.

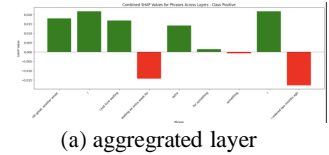

(a) aggregated layer

Figure 13: Proposed Approach for Sarcasm.

The proposed approach overcomes the baseline's limitations by providing a comprehensive, context-aware, and layered explanation of the model's sentiment analysis process.

## Conclusion

The aim of this research was to improve the explainability of large language models in sentiment analysis by decomposing the model into key components, including encoder, decoder, embedding layer, and attention, and applying SHAP to each layer for a detailed understanding of their contributions. This layer-by-layer approach addresses the limitations of whole model explainability methods by uncovering how semantic features and contextual relationships evolve across layers. The experiments showed that analyzing SHAP values at the embedding and encoder layers provides precise insights into how individual phrases influence sentiment predictions. Notably, the proposed approach effectively identified nuances like sarcasm, a challenge in sentiment analysis, by revealing how it impacts sentiment attribution across different layers. This capability enhances the transparency and trustworthiness of sentiment analysis models, making them more suitable for high stakes applications. The findings emphasize the importance of component wise SHAP analysis for improving transparency and reliability. This method not only explains what drives a model's predictions but also how sentiment evolves through layers, offering deeper interpretability. Future work can focus on optimizing this framework for larger models and extending it to multi modal architectures to broaden its scope while maintaining its emphasis on explainability.

# References

Abnar, S.; and Zuidema, W. 2021. Explaining transformers using hierarchical attention attribution. *Proceedings of EMNLP*, 11–20.

Arras, L.; Montavon, G.; Müller, K.-R.; and Samek, W. 2020. Explaining and interpreting LSTMs using layer-wise relevance propagation. *IEEE Transactions on Neural Networks and Learning Systems*, 31(11): 4786–4797.

Bahdanau, D.; Cho, K.; and Bengio, Y. 2015. Neural machine translation by jointly learning to align and translate. In *International Conference on Learning Representations (ICLR)*.

Barnhart, J.; and Balakumar, K. 2020. SHAP Values for Deep Transformer Models: Insights into BERT for Sentiment Analysis. In *Proceedings of the 2020 Conference on Empirical Methods in Natural Language Processing*.

Chefer, H.; Gur, S.; and Wolf, L. 2020. Transformer interpretability beyond attention visualization. *Proceedings of CVPR*, 6221–6230.

Chen, X.; Zhang, H.; and Sun, K. 2023. TransSHAP: Transformer-based SHAP explanations for interpretability in NLP models. In *Proceedings of EMNLP*, 1800–1812.

Devlin, J.; Chang, M.-W.; Lee, K.; and Toutanova, K. 2019. BERT: Pre-training of deep bidirectional transformers for language understanding. In *Proceedings of NAACL-HLT*, 4171–4186.

Doshi-Velez, F.; and Kim, B. 2017a. Towards a rigorous science of interpretable machine learning. *arXiv preprint arXiv:1702.08608*.

Doshi-Velez, F.; and Kim, B. 2017b. Towards a rigorous science of interpretable machine learning. In *arXiv preprint arXiv:1702.08608*.

Gunning, D.; and Aha, D. W. 2019. Explainable artificial intelligence (XAI): Concepts, taxonomies, opportunities, and challenges toward responsible AI. *Proceedings of AAAI*, 33: 9783–9787.

Jain, S.; and Wallace, B. C. 2019. Attention is not explanation. *Proceedings of NAACL-HLT*, 3543–3556.

Kokalj, E.; Škrlj, B.; Lavrač, N.; Pollak, S.; and Robnik-Šikonja, M. 2021. BERT meets Shapley: Extending SHAP explanations to transformer-based classifiers. *Journal of Artificial Intelligence Research*, 72: 215–243.

Lei, T.; Barzilay, R.; and Jaakkola, T. 2016. Rationalizing neural predictions. *Proceedings of EMNLP*, 107–117.

Lundberg, S. M.; and Lee, S.-I. 2017. A unified approach to interpreting model predictions. *Advances in Neural Information Processing Systems*, 30.

Lundberg, S. M.; and Lee, S.-I. 2020. Local explainability of machine learning models using Shapley values. In *Proceedings of the 2020 Conference on Fairness, Accountability, and Transparency*, 277–288.

Maas, A. L.; Daly, R. E.; Pham, P. T.; Huang, D.; Ng, A. Y.; and Potts, C. 2011. Learning Word Vectors for Sentiment Analysis. In *Proceedings of the 49th Annual Meeting of the Association for Computational Linguistics: Human Language Technologies*, 142–150. Portland, Oregon, USA: Association for Computational Linguistics.

Mackova, K.; Wang, R.; and Mitra, T. 2021. SHAP for sentiment analysis in the healthcare domain. In *Proceedings of the International Conference on Computational Semantics*, 10–18.

Macková, P.; Šimko, M.; and Bieliková, M. 2021. Extended SHAP for Understanding Transformer Models: Insights into Attention Mechanisms. In *Proceedings of the 2021 Conference on Empirical Methods in Natural Language Processing*.

Narayan, A.; and Verma, S. 2021. Application of LIME to Explain Transformer-Based Models in Healthcare. In *Proceedings of the 2021 Conference on Neural Information Processing Systems (NeurIPS)*.

Pang, B.; and Lee, L. 2008. Opinion mining and sentiment analysis. *Foundations and Trends in Information Retrieval*, 2(1–2): 1–135.

Radford, A.; Narasimhan, K.; Salimans, T.; and Sutskever, I. 2018. Improving language understanding by generative pre-training. *OpenAI Blog*, 1(8).

Ribeiro, M. T.; Singh, S.; and Guestrin, C. 2016. "Why should I trust you?" Explaining the predictions of any classifier. In *Proceedings of the 22nd ACM SIGKDD International Conference on Knowledge Discovery and Data Mining*, 1135–1144. ACM.

Rudin, C. 2019. Stop explaining black box machine learning models for high-stakes decisions and use interpretable models instead. *Nature Machine Intelligence*, 1(5): 206–215.

Socher, R.; Perelygin, A.; Wu, J.; Chuang, J.; Manning, C. D.; Ng, A. Y.; and Potts, C. 2013. Recursive deep models for semantic compositionality over a sentiment treebank. In *Proceedings of the 2013 conference on empirical methods in natural language processing (EMNLP)*, 1631–1642.

Sun, C.; Huang, L.; and Qiu, X. 2019. Utilizing BERT for aspect-based sentiment analysis via constructing auxiliary sentence. In *Proceedings of NAACL-HLT*, 380–385.

Sundararajan, M.; Taly, A.; and Yan, Q. 2017. Axiomatic attribution for deep networks. *Proceedings of ICML*, 70: 3312–3320.

Zhang, L.; Wang, S.; and Liu, B. 2021. Sentiment analysis: A survey. *IEEE Transactions on Affective Computing*, 12(2): 264–283.

Zhao, K.; Chen, Y.; and Yang, X. 2024. Analyzing multi-head self-attention for model interpretability. In *Proceedings of EMNLP 2024*.